\documentclass[letterpaper]{article} 
\usepackage{aaai2026}  
\usepackage{times}  
\usepackage{helvet}  
\usepackage{courier}  
\usepackage[hyphens]{url}  
\usepackage{graphicx} 
\usepackage{natbib}  
\usepackage{caption} 
\usepackage{algorithm}
\usepackage{algorithmic}

\usepackage{subfig}
\usepackage{comment}
\usepackage{amsfonts}
\usepackage{newfloat}
\usepackage{listings}
\DeclareCaptionStyle{ruled}{labelfont=normalfont,labelsep=colon,strut=off} 
\floatstyle{ruled}
\newfloat{listing}{tb}{lst}{}
\floatname{listing}{Listing}
\title{HN-MVTS: HyperNetwork-based Multivariate Time Series Forecasting}
\author{Andrey Savchenko\textsuperscript{\rm 1,\rm 2}, Oleg Kachan\textsuperscript{\rm 1, \rm 2}\\
}
\affiliations {
    \textsuperscript{\rm 1}Sber AI Lab, Moscow, Russia\\
    \textsuperscript{\rm 2}HSE University, Moscow, Russia \\
    avsavchenko@hse.ru, okachan@hse.ru
}

\begin{document}

\maketitle

\begin{abstract}
Accurate forecasting of multivariate time series data remains a formidable challenge, particularly due to the growing complexity of temporal dependencies in real-world scenarios. While neural network-based models have achieved notable success in this domain, complex channel-dependent models often suffer from performance degradation compared to channel-independent models that do not consider the relationship between components but provide high robustness due to small capacity. In this work, we propose HN-MVTS, a novel architecture that integrates the hypernetwork-based generative prior with an arbitrary neural network forecasting model. The input of this hypernetwork is a learnable embedding matrix of time series components. To restrict the number of new parameters, the hypernetwork learns to generate the weights of the last layer of the target forecasting networks, serving as a data-adaptive regularizer that improves generalization and long-range predictive accuracy. The hypernetwork is only used during training, so it does not increase the inference time compared to the base forecasting model. Extensive experiments on eight benchmark datasets demonstrate that application of HN-MVTS to the state-of-the-art models (DLinear, PatchTST, TSMixer, etc.) typically improves their performance. Our findings suggest that hypernetwork-driven parameterization offers a promising direction for enhancing existing forecasting techniques in complex scenarios.
\end{abstract}

\begin{links}
     \link{Code}{https://github.com/av-savchenko/HN-MVTS}
\end{links}

\section{Introduction}
Multivariate time series (MVTS) forecasting~\cite{ijcai2025p1266,mendis2024multivariate} is critical to many real-world applications, including energy demand prediction, traffic management, financial forecasting, weather modeling, and health monitoring. In such scenarios, multiple interrelated time series, often referred to as channels or components, must be predicted simultaneously. Effective forecasting, therefore, depends not only on modeling temporal patterns within each series but also on capturing correlations across components.

Advancements in deep learning have led to powerful forecasting models, ranging from linear baselines~\cite{Zeng2023} to architectures based on convolutional networks~\cite{Luo2024}, multilayer perceptrons (MLPs)~\cite{Chen2023}, and Transformers~\cite{Nie2023}. Surprisingly, recent work has shown that channel-independent (CI) models, which treat each time series channel separately, often outperform channel-dependent (CD) approaches that attempt to model all channels jointly~\cite{Han2024,Peiwen2023}. CI models reduce model complexity, but sacrifice the ability to leverage valuable inter-channel relationships, potentially limiting their accuracy in highly correlated settings. 

On the other hand, CD models explicitly exploit inter-channel dependencies. Indeed, the improvement over baseline is often achieved by advanced modeling of temporal \cite{Nie2023} and channel \cite{Liu24} similarities, or both \cite{Wu2023}. However, CD models are often known to be less scalable due to their requirement for larger datasets~\cite{Peiwen2023}. Designing models that can balance the robustness of CI approaches with the expressiveness of CD strategies remains a key open challenge in MVTS forecasting~\cite{Wang2025,zhao2024rethinking}. The difficulty lies in balancing the leverage of inter-channel correlations and maintaining robustness and scalability across diverse datasets. For example, a local model with individual parameters for each channel results in increased parameters and does not capture the channels' similarities~\cite{montero2021localvsglobal}.

\begin{figure*}[t]
\centering
\includegraphics[width=1.0\textwidth]{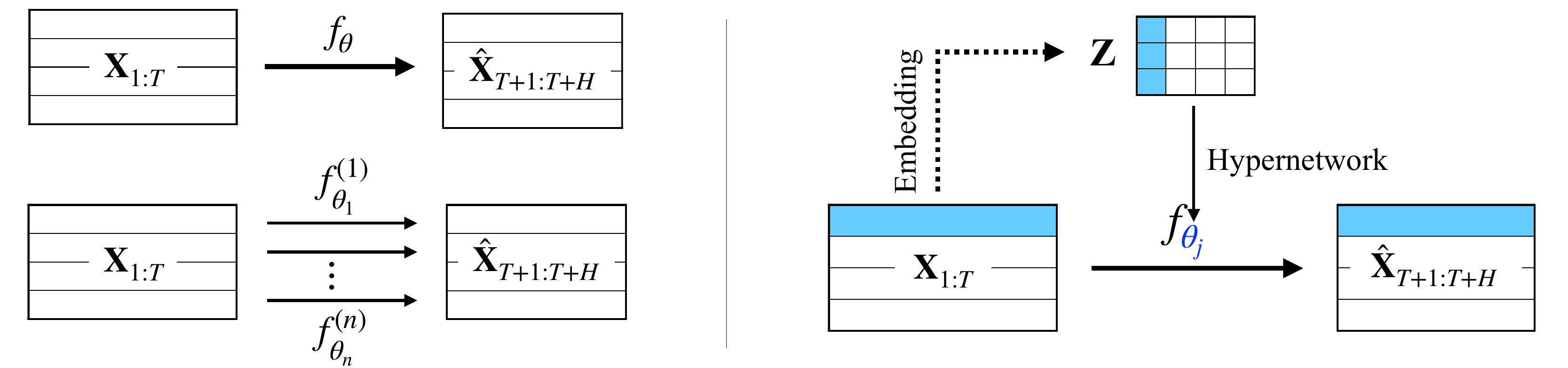}
\caption{\textbf{Top left: }Channel-dependent, \textbf{Bottom left:} channel-independent, and \textbf{Right:} proposed HN-MVTS forecasting approach, where a hypernetwork outputs parameters of the forecasting model's last layer, taking a channel embedding as an input.}\label{fig:forecasting_strategies}
\end{figure*}

To address this, we propose HN-MVTS (Fig.~\ref{fig:forecasting_strategies}), a novel architecture that bridges the gap between channel-independent and channel-dependent modeling by using a hypernetwork 
to generate channel-specific forecasting parameters. Specifically, a small MLP-based hypernetwork receives a learnable embedding for each channel and outputs the weights of the final prediction layer for that channel. This setup allows similar channels to share information through their embeddings while preserving the modularity and efficiency of CI models. A key strength of HN-MVTS is its architectural flexibility: it can be seamlessly integrated with a wide range of neural network-based forecasting models, including linear baselines, MLPs, ConvNets, and Transformer variants. By modifying only the final prediction layer, HN-MVTS enhances the forecasting performance of existing architectures with minimal overhead in training time and the number of learnable parameters and without any impact on the inference time. We demonstrate the effectiveness of HN-MVTS through comprehensive experiments on multiple benchmark datasets. Our approach typically substantially improves the performance of several state-of-the-art models, such as DLinear, PatchTST, and TSMixer, on multiple widely used datasets (e.g., ECL, ETTh), highlighting its robustness, adaptability, and ability to generalize across diverse settings.

\section{Related Works}
\textbf{Models.}
Nowadays, the dominant architecture in MVTS forecasting is the Transformer. Modern papers illustrate an expanding design space where model efficacy increasingly depends on architectural alignment with the unique characteristics of time series~\cite{su2025systematic,wang2024deep}. Inverse Transformer (iTransformer)~\cite{Liu24} processes entire time points as variate tokens rather than temporal tokens, enabling better capture of multivariate correlations while retaining Transformer components. It addresses key limitations of conventional temporal tokenization, particularly when handling large lookback windows and diverse variate relationships. 
PatchTST~\cite{Nie2023} demonstrates how patching from computer vision can adapt Transformers for time series through subseries-level embeddings and channel independence. 

However, recent works have introduced various models that challenge traditional architectures by focusing on improved temporal and cross-variable dependencies. For example, the DLinear architecture~\cite{Zeng2023} has emerged as a compelling alternative by decomposing series into trend and seasonality components through linear layers, achieving superior performance over complex Transformer variants, such as FEDformer~\cite{zhou2022fedformer}, on key benchmarks while maintaining computational efficiency. This simplicity-first approach sparked reevaluations of model design priorities, with ModernTCN~\cite{Luo2024} further advancing non-Transformer paradigms through modernized temporal convolutional networks that expand effective receptive fields while maintaining computational efficiency. TSMixer~\cite{Chen2023}, TimesMixer~\cite{wangtimemixer}, and HDMixer~\cite{huang2024hdmixer} exemplify this through an all-MLP (multi-layer perceptron) architecture that integrates historical data, future known inputs, and static covariates using conditional feature mixing layers, demonstrating competitive performance against Transformers. 

Finally, techniques based on graph neural networks~\cite{makarov2022temporal}, such as graph inter-series models like MSGNet~\cite{cai2024msgnet} and FourierGNN~\cite{yi2023fouriergnn}, or ForecastGrapher~\cite{cai2024forecastgrapher}, treat each series as a node and learn a graph; however, their scalability is usually relatively poor.

\textbf{Channel-dependent vs channel-independent models.}
MVTS forecasting strategies are broadly categorized into CI and CD~\cite{montero2021localvsglobal}. The CI mode~\cite{Han2024}, which includes both global (single model applied to all univariate series) and local (dedicated models per series) options, treats each time series channel as independent, ignoring cross-channel correlations. In contrast, the CD strategy employs a single model that takes all components as input and forecasts all components simultaneously, explicitly modeling inter-channel dependencies~\cite{hertel2023transformer}. The CI-CD performance gap stems from fundamental trade-offs between model capacity and robustness. CD methods theoretically possess a higher capacity for modeling complex channel interactions, but struggle with real-world challenges, such as limited training data. CI strategies overcome these issues by using larger effective datasets (through channel-wise replication) and avoiding error propagation from correlated channels. Hence, empirical studies~\cite{hertel2023transformer} demonstrate that CI methods outperform CD approaches across various domains, including energy load forecasting, financial data analysis, and epidemiological modeling. \cite{Han2024} benchmarks global univariate versus multivariate modeling for linear, autoregressive, MLP, gradient boosting, and transformer-based models. They show that CI (univariate) models generally significantly outperform multivariate ones, though most transformer-based MVTS forecasting models are trained using the multivariate strategy. Global CI methods achieve parameter efficiency comparable to local models while maintaining competitive accuracy~\cite{montero2021localvsglobal}. Indeed, the top linear model (DLinear)~\cite{Zeng2023} trained using the global univariate approach outperforms multivariate-trained transformers. Similarly, \cite{hertel2023transformer} reports that global CI Transformers significantly reduce forecasting errors compared to CD counterparts in smart grid applications. 
The success of CI highlights the importance of robustness in real-world forecasting~\cite{ashouri2022interactive}. Nevertheless, CD approaches retain promise for domains with strong inter-channel correlations, provided they incorporate mechanisms like temporal clustering or residual regularization~\cite{Han2024}. For example, DUET~\cite{qiu2024duet} introduces dual temporal-channel clustering to enhance CD models, showing that structured dependency modeling combined with robustness mechanisms can match CI performance in scenarios with clear channel groupings. PatchTST~\cite{Nie2023} processes time series into patched ``words'' using CI Transformers to preserve univariate patterns while enabling cross-variate attention. \cite{chen2024clustering} observes the boost in forecasting performance for multivariate models for correlated time series. These developments underscore the need for context-aware strategy selection, balancing the CI-CD trade-off against dataset characteristics and operational constraints. Similar challenges are addressed in COSCI-GAN~\cite{seyfi2022generating} for another task, generative modeling, by coordinating single-channel generators from a common noise source and utilizing a central discriminator to enforce inter-channel relationships. However, it does not employ an embedding-based parameter generation approach.

\textbf{Hypernetworks.}
Hypernetworks, neural networks that generate weights for another neural network (the target network), have emerged as a powerful approach for various machine learning tasks~\cite{ha2016hypernetworks}, including time series forecasting. 
The foundational work~\cite{duan2023combating} introduced Hyper Time Series Forecasting, which jointly learns time-varying distributions and corresponding forecasting models, enabling more accurate predictions when underlying data distributions change over time. Recent advances include HyperGPA~\cite{lee2022time}, a hypernetwork that generates optimal model parameters for each time period through computational graph structures. It demonstrates the effectiveness in forecasting scenarios where future data patterns differ from historical observations. 
Finally, LPCNet employs a hypernetwork-based framework that dynamically updates neural network parameters in response to changing input distributions~\cite{liu2024hypernetwork}. This adaptive parameter generation automatically enables the model to adjust to varying temporal patterns with acceptable computational complexity. 
These advancements highlight the ability of hypernetworks to model complex temporal dependencies while adapting to evolving data distributions.

Another research direction is the usage of a hypernetwork for meta-learning to handle multiple tasks. Hyper-Trees~\cite{marz2024forecasting} generate parameters that adapt to local characteristics when applied to individual series, enabling effective handling of heterogeneous time series. Recent work~\cite{stanvek2025designing} explored the hypernetwork's meta-learning architectures, which are capable of constructing optimal parametric models for families of similar data-generating processes, achieving competitive performance in forecasting challenges. The HyperTime framework~\cite{fons2022hypertime} provides an accurate and resolution-independent encoding of time series data, facilitating accurate reconstruction and interpolation. It was effective for data augmentation, achieving competitive performance against state-of-the-art time series augmentation methods.

Thus, hypernetworks show promise for limited observations where nonparametric methods fail. However, their applications are still restricted to managing non-stationary time series, data augmentation, or training when multiple realizations of similar but not identical time series are available. To the best of our knowledge, they have not been applied to improve the performance metrics, such as mean squared error (MSE), of state-of-the-art models for MVTS forecasting. This paper fills this gap by utilizing hypernetworks to leverage the benefits of both CI and CD approaches.

\section{Proposed Approach}

\begin{figure*}[t]
  \centering
  \includegraphics[width=\textwidth]{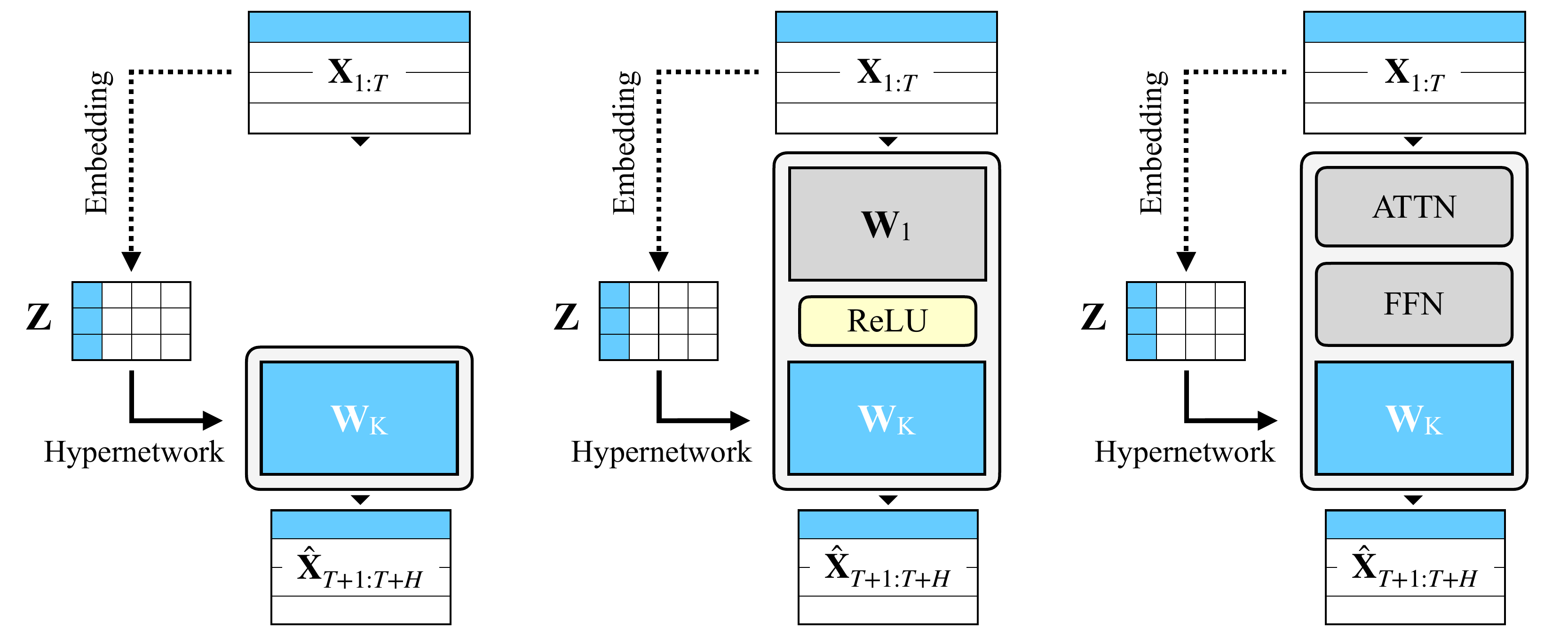}
  \caption{Examples of adding our HN-MVTS to various forecasting models. \textbf{Left:} Linear model. \textbf{Center:} Fully-connected neural network. \textbf{Right:} Transformer model with attention (ATTN) and feed-forward (FFN) layers.}\label{fig:hypernet_architectures}
\end{figure*}

\paragraph{Problem statement.} 
Let the history be $\mathbf{X}_{1:T} \in \mathbb{R}^{N \times T}$, where $N \ge 1$ is the number of channels, and $T$ is the length of the lookback window. The multivariate time series forecasting task is to predict $\mathbf{X}_{T+1:T+H} \in \mathbb{R}^{N \times H}$, where $H \ge 1$ is the prediction horizon.


We consider a supervised learning scenario, where time series $\mathbf{X}_{1:t} \in \mathbb{R}^{N \times t}$ with duration $t \gg T$ are available. It is typically used to create the training dataset of pairs $\{(\mathbf{X}_{i:T+i}, \mathbf{X}_{T+1+i:T+H+i}) \}_{i=0}^M$, where the dataset size $M=t-(T+H)$. This dataset is used to learn the weights $\mathbf{\theta}$ of the model $f_\theta$ that predicts future $H$ steps for each of $N$ channels: 
\begin{equation}\label{eq:base_model}
    \mathbf{\hat{X}}_{T+1+i:T+H+i} = f_\theta(\mathbf{X}_{i:T+i}).
\end{equation}

The MSE loss function is typically minimized to obtain the optimal weights $\theta^*$. 

In this paper, we focus on neural network models $f_\theta$ with $K$ layers, and \textit{assume} that its last ($K$-th) layer linearly transforms an input hidden state $[\mathbf{h}^{(1)}, \dots, \mathbf{h}^{(N)}]$, where $\mathbf{h}^{(n)}\in \mathbb{R}^D$ with hidden dimension $D$ into the final output $\mathbf{\hat{X}}=[\hat{\mathbf{x}}^{(1)}, \dots, \hat{\mathbf{x}}^{(N)}] \in \mathbb{R}^{N \times H}$ using the matrix product for each channel $n=1, \dots, N$:
\begin{equation}\label{eq:last_layer}
    \hat{\mathbf{x}}_n= \textbf{W}_K^{(n)} \cdot \mathbf{h}^{(n)} ,
\end{equation}
where $\textbf{W}_K^{(n)} \in \mathbb{R}^{H \times D}$ 
is a weight matrix for the $n$-th channel of the last layer.
 

\paragraph{Hypernetworks.}

Our core idea is to bridge the gap between CI and CD forecasting models by using a hypernetwork to generate channel-specific parameters based on learnable component embeddings. In contrast to conventional models that use a fixed set of parameters across all time series components, we dynamically adapt the final prediction layer for each channel, enabling personalized forecasting behavior while preserving parameter efficiency.

Let $f_{\theta_1, \dots, \theta_K}: \mathcal{X} \rightarrow \mathcal{Y}$ be a base neural network, parameterized by $\mathbf{\theta}=\{\theta_k\}_{k=1}^K$, where $\theta_{k}$ are the parameters of the $k$-th layer. A hypernetwork $h_\phi: \mathcal{Z} \rightarrow \Theta$ predicts parameters of the base network $f$ given a task embedding $\mathbf{z}$, resulting a network $f_{h_\phi}: \mathcal{X} \times \mathcal{Z} \rightarrow \mathcal{Y}$~\cite{ha2016hypernetworks}. 


The main idea of this paper is to parametrize the weights $\theta_K=\textbf{W}_K = [\textbf{W}_K^{(1)}, \dots, \textbf{W}_K^{(N)}]\in \mathbb{R}^{N \times H \times D}$ of the last (prediction) layer (\ref{eq:last_layer}) with a partial hypernetwork. In particular, each $n$-th component of a time series ($n \in \{1,2,\dots,N\}$) is associated with a $d$-dimensional embedding vector $\mathbf{z}^{(n)} \in \mathbb{R}^d$, and the matrix $\textbf{Z}=[\mathbf{z}^{(1)}, \dots, \mathbf{z}^{(N)}]\in \mathbb{R}^{N \times d}$ is fed into a hypernetwork:
\begin{equation}\label{eq:hyper}
\textbf{W}_K=h_\phi(\textbf{Z}). 
\end{equation}

In this paper, we implement a hypernetwork (\ref{eq:hyper}) as a MLP with one hidden layer that that takes a learnable embedding vector for each time series component $\mathbf{z}^{(n)}$ (treating the number of channels $N$ as the batch size) and outputs the corresponding weights for the final prediction layer (\ref{eq:last_layer}) of an arbitrary base forecasting model. In the simplest case of an MLP without hidden layers, the weights of the base model are obtained using a linear transform:
\begin{equation}\label{eq:linearhyper}
\textbf{W}_K^{(n)}=\textbf{W}_\phi^{(n)} \cdot \mathbf{z}^{(n)}, 
\end{equation}
where  $\textbf{W}_\phi^{(n)} \in \mathbb{R}^{H \times D \times d}$ 
is the $n$-th component of the weights $\textbf{W}_\phi^{(n)}=[\textbf{W}_\phi^{(1)}, \cdots, \textbf{W}_\phi^{(N)}] \in \mathbb{R}^{N \times H \times D \times d}$ of the hypernetwork.

The central contribution of our hypernetwork-based approach (HN-MVTS) lies in utilizing a hypernetwork to generate part of the parameters of a forecasting model in a computationally efficient yet scalable and data-adaptive manner. In traditional MVTS models, all output channels typically share the same output projection layer (CD), or each channel has its own separate model (CI). Our approach aims to combine both benefits by introducing a shared neural forecasting model, while tailoring its final layer to each channel via a learnable hypernetwork. The difference between CD, CI, and our HN-MVTS is shown in Fig.~\ref{fig:forecasting_strategies}. The former is the most common model, with a single transform $f_\theta$ of $N \times T$ input data into $N \times H$ outputs. The CI approach learns $N$ separate models $f_{\theta_1}^{(1)}, \dots, f_{\theta_1}^{(N)}$, and processes each channel independently. Our HN-MVTS saves the advantages of both types of models. If some components $j_1$ and $j_2$ of time series are similar, their embeddings will be close to each other $\mathbf{z}_{j_1} \approx \mathbf{z}_{j_2}$, and, hence, the training data for the $j_1$ component will have more influence to learn the weights for the $j_2$ component, and vice versa. In the ideal case of clustered embeddings  $\mathbf{z}_{j_1} = \mathbf{z}_{j_2}$, a single (global) model~\cite{montero2021localvsglobal} is trained for the union of these components. In contrast, if the embeddings of  $j_1$ and $j_2$ channels are significantly different, a CI mode is activated, in which the forecasting parameters are trained independently. This mechanism allows the model to adaptively interpolate between CD and CI behavior based on the learned embedding, without manual architecture changes or channel grouping. 


Some examples of adding our HN-MVTS to various architectures are shown in Fig.~\ref{fig:hypernet_architectures}. The above-mentioned motivation of mixing CD and CI models is especially clear for a forecasting model with one linear layer (left part of this figure), where the columns of the weight matrix $W$ are learned based on the similarity between embeddings $\mathbf{z}$ of time series components. However, our approach can be easily applied to MLP (Fig.~\ref{fig:hypernet_architectures}b) or more complex architectures such as transformers (Fig.~\ref{fig:hypernet_architectures}c), as only the weights of the last prediction layer are defined by the hypernetwork. It may be used for all layers, but the number of trained parameters will be significantly greater, making it more challenging to train the model. A practitioner can control the number of excess parameters by managing the complexity of the hypernetwork. 

In all experiments, we use the following straightforward implementation of HN-MVTS. First, we initialize embeddings $\mathbf{z}$ from the Pearson product-moment correlation coefficients across channels~\cite{nguyen2024learning}, which are projected onto the principal components of dimension $d$ computed using the training-only splits. Our experiments indicate that alternative (random) initialization results in slightly higher MSE. Next, we attach a simple MLP (\ref{eq:linearhyper}) and learn its weights and embeddings $\mathbf{z}$ from given data together with the base model (\ref{eq:base_model}) by minimizing MSE. Our simplest implementation (\ref{eq:linearhyper}) adds only $N \cdot H \cdot D \cdot d$ 
parameters to learn (with addition of $N \cdot d$ if embeddings $Z$ are learnable). As a result, the number of parameters can be even less than that of the CI approach, which applies the base model separately with its parameters for each channel. As the number of learnable weights is reasonably small, the training time is only slightly higher than that of the base forecasting model.

Moreover, since the trained hypernetwork outputs fixed weights that do not depend on the specific input time series, we can compute them only once after the training procedure is complete. Hence, we remove a hypernetwork and copy weights $\textbf{W}_K^{(n)}$ (\ref{eq:linearhyper}) into the last linear layer of the base model (\ref{eq:last_layer}). As a result, the inference time won't be affected (though deployment overheads, e.g., I/O, memory layout, may still cause minor variation), making our HN-MVTS as fast as the base model.

Thus, the proposed approach uses a learnable component embedding combined with a hypernetwork to produce channel-specific final-layer weights as a data-adaptive regularizer, integrates seamlessly with diverse backbone architectures, and preserves inference ~\cite{savchenko2016search}. These points with empirical gains discussed in the next section are a substantive contribution beyond the straightforward reapplication of existing hypernetwork's ideas.

%






\section{Experimental setup}\label{subsec:exper_setup}

\paragraph{Baseline models.}
We evaluated the proposed HN-MVTS with contemporary multivariate time series forecasting models of different architectures: 1) Linear model: DLinear \cite{Zeng2023}, 2) MLP-based model: TSMixer \cite{Chen2023}, 3) Convolutional neural network-based model: ModernTCN \cite{Luo2024}, and 4) Transformer-based models: PatchTST \cite{Nie2023} and inverse Transformer \cite{Liu24}. The source code for all models was taken from the original repositories of their authors.

\paragraph{Datasets.}
We leverage eight publicly available open datasets that are conventionally used in the MVTS literature~\cite{10726722}, covering energy, traffic, and weather forecasting applications: \texttt{ECL} (UCI Electricity Consuming Load) with electricity consumption of hundreds of points/clients; 15-minute-level datasets (\texttt{ETTm1, ETTm2}) from the Electricity Transformer Temperature database~\cite{zhou2021informer} with one oil and six load features of electricity transformers from 2016 to 2018; \texttt{Weather} (Max-Planck-Institute Weather Dataset for Long-term Time Series Forecasting) with weather indicators recorded every 10 min for 2020, and Traffic Flow from Performance Measurement System (\texttt{PEMS03}, \texttt{PEMS04}, \texttt{PEMS07}, \texttt{PEMS08}) collected by California Transportation Agencies (CalTrans) every 30 seconds and aggregated to 5 minutes. 


Each dataset exhibits diverse temporal and structural properties, allowing us to assess the robustness and generalizability of our method. Their statistics are summarized in Table \ref{table_datasets}. We split the time series into train, validation, and test sets using a 7:2:1 ratio. For the \texttt{ETTm1} and \texttt{ETTm2} datasets, following standard testing protocols, we use a $6:2:2$ split and take only the first 57600 timesteps. The validation set was used to obtain the best checkpoint of the model, and the results on the test set are reported.

\begin{table}[t]
\begin{center}
\resizebox{\columnwidth}{!}{
\begin{tabular}{l|l|l|l|l}
Dataset       & Timesteps $t$ & Channels $N$ & Split ratio & Granularity \\ \hline
ECL     & 26304      & 321        & 7:2:1       & 1h          \\
ETTm1   & 57600      & 7          & 6:2:2       & 15m         \\
ETTm2   & 57600      & 7          & 6:2:2       & 15m         \\
PEMS03  & 26208      & 358        & 7:2:1       & 5m          \\
PEMS04  & 16992      & 307        & 7:2:1       & 5m          \\
PEMS07  & 28224      & 883        & 7:2:1       & 5m          \\
PEMS08  & 17856      & 170        & 7:2:1       & 5m          \\
Weather & 52696      & 21         & 7:2:1       & 10m         \\ \end{tabular}
}
\caption{Datasets properties.}
\label{table_datasets}

\end{center}
\end{table}

\begin{table*}[h!]
\begin{center}
\resizebox{\textwidth}{!}{
\begin{tabular}{l|l|ll|ll|ll|ll|ll}
Dataset & $H$ & DLinear & +HN-MVTS & TSMixer& +HN-MVTS & ModernTCN& +HN-MVTS   & PatchTST& +HN-MVTS & iTransformer& +HN-MVTS \\
\hline
& 48 & 0.1255 & \textbf{0.1184} & 0.1377 & \textbf{0.1220}& 0.1270& \textbf{0.1191} & 0.1162& 0.1157& 0.1091& \textbf{0.1076}\\
ECL & 96 & 0.1409& \textbf{0.1347}& 0.1748& \textbf{0.1421}& 0.1516& \textbf{0.1383}& 0.1328& \textbf{0.1318} & 0.1309& \textbf{0.1297}\\
 & 192 & 0.1577& \textbf{0.1533} & 0.1972 & \textbf{0.1666} & 0.1783 & \textbf{0.1708} & 0.1523 & 0.1532 & 0.1575 & \textbf{0.1536} \\
 & 336 & 0.1720 & \textbf{0.1676} & 0.1986 & \textbf{0.1747} & 0.1804 & \textbf{0.1750} & 0.1669 & 0.1681 & 0.1690& \textbf{0.1637} \\
\hline
& 48 & 0.2705& \textbf{0.2610}& \textbf{0.2669}& 0.2749& 0.2615& \textbf{0.2597}& \textbf{0.2528}& 0.2603& 0.2871& 0.2863\\
ETTm1 & 96 & 0.3019& \textbf{0.2899}& 0.3002& 0.2969& 0.2929& \textbf{0.2899}& 0.2899& 0.2898& \textbf{0.3034}& 0.3171\\
 & 192 & 0.3364& \textbf{0.3306}& 0.3405& \textbf{0.3361}& 0.3455& \textbf{0.3397}& 0.3295& 0.3313& \textbf{0.3386}& 0.3579\\
 & 336 & 0.3717& 0.3721& 0.3820& \textbf{0.3759}& 0.3824& \textbf{0.3760}& 0.3657& 0.3669& \textbf{0.3767}& 0.3946\\
\hline
& 48 & 0.1279& \textbf{0.1246}& 0.1373& \textbf{0.1326}& 0.1251& \textbf{0.1230}& 0.1302& 0.1310& 0.1363& 0.1376\\
ETTm2 & 96 & 0.1641& \textbf{0.1626}& 0.1805& \textbf{0.1749}& 0.1641& \textbf{0.1606}& 0.1696& 0.1703& 0.1823& 0.1832\\
 & 192 & 0.2183& \textbf{0.2165}& 0.2375& \textbf{0.2308}& 0.2285& 0.2297& 0.2339& 0.2357& 0.2538& \textbf{0.2448}\\
 & 336 & 0.2746& \textbf{0.2730}& 0.2906& \textbf{0.2848}& \textbf{0.2857}& 0.3075& 0.2885& 0.2903& 0.3055& 0.3055\\
\hline
& 48 & 0.1513& \textbf{0.1490}& 0.1077& 0.1073& 0.1313& \textbf{0.1178}& 0.1143& \textbf{0.1071}& 0.0898& \textbf{0.0875}\\
PEMS03 & 96 & 0.1899& \textbf{0.1871}& 0.1369& 0.1376& \textbf{0.1828}& 0.2063& 0.1461& \textbf{0.1364}& 0.1118& \textbf{0.1087}\\
 & 192 & 0.2096& \textbf{0.2065}& 0.1537& 0.1554& 0.2004& \textbf{0.1629}& 0.1773& \textbf{0.1719}& 0.1307& \textbf{0.1223}\\
 & 336 & 0.2265& \textbf{0.2227}& 0.1654& 0.1645& 0.2505& \textbf{0.2135}& 0.1899& \textbf{0.1862}& 0.1504& \textbf{0.1481}\\
\hline
& 48 & 0.1639& \textbf{0.1572}& 0.1185& \textbf{0.1101}& 0.1048& 0.1056& 0.1306& \textbf{0.1130}& 0.0960& \textbf{0.0878}\\
PEMS04 & 96 & 0.2016& \textbf{0.1943}& 0.1312& \textbf{0.1270}& 0.1744& \textbf{0.1682}& 0.1627& \textbf{0.1461}& 0.1136& \textbf{0.1030}\\
 & 192 & 0.2208& \textbf{0.2125}& 0.1566& \textbf{0.1395}& 0.1874& \textbf{0.1662}& 0.1873& \textbf{0.1638}& 0.1267& \textbf{0.1189}\\
 & 336 & 0.2444& \textbf{0.2390}& 0.1760& \textbf{0.1519}& 0.1985& \textbf{0.1786}& 0.2005& \textbf{0.1741}& 0.1533& \textbf{0.1333}\\
\hline
& 48 & 0.1479& \textbf{0.1411}& 0.0959& \textbf{0.0904}& 0.1328& \textbf{0.1070}& 0.0992& \textbf{0.0888}& 0.0681& \textbf{0.0637}\\
PEMS07 & 96 & 0.1871& \textbf{0.1766}& 0.1117& \textbf{0.1062}& \textbf{0.1253}& 0.1474& 0.1240& \textbf{0.1092}& 0.0790& \textbf{0.0721}\\
 & 192 & 0.2057& \textbf{0.1994}& 0.1308& \textbf{0.1212}& \textbf{0.1399}& 0.1489& 0.1478& \textbf{0.1286}& 0.0908& \textbf{0.0818}\\
 & 336 & 0.2287& \textbf{0.2200}& 0.1533& \textbf{0.1326}& \textbf{0.1442}& 0.1545& 0.1619& \textbf{0.1415}& 0.1030& \textbf{0.0928}\\
\hline
& 48 & 0.1889& \textbf{0.1651}& 0.1120& \textbf{0.1027}& 0.1268& 0.1250& 0.1221& \textbf{0.1052}& 0.0870& \textbf{0.0799}\\
PEMS08 & 96 & 0.2618& \textbf{0.2257}& 0.1482& \textbf{0.1268}& \textbf{0.2284}& 0.2356& 0.1562& \textbf{0.1348}& 0.1113& \textbf{0.0957}\\
 & 192 & 0.3067& \textbf{0.2780}& 0.2217& \textbf{0.1884}& 0.2327& \textbf{0.1894}& 0.2128& \textbf{0.1964}& 0.1459& \textbf{0.1187}\\
 & 336 & 0.3273& \textbf{0.3252}& 0.2496& 0.2477& 0.2694& \textbf{0.2640}& 0.2343& \textbf{0.2231}& 0.1770& \textbf{0.1769}\\
\hline
& 48 & 0.1369& \textbf{0.1115}& 0.1168& 0.1162& 0.1151& \textbf{0.1127}& 0.1136& 0.1143& 0.1235& \textbf{0.1195}\\
Weather & 96 & 0.1733& \textbf{0.1425}& 0.1490& \textbf{0.1475}& 0.1480& \textbf{0.1429}& \textbf{0.1448}& 0.1480& 0.1559& \textbf{0.1505}\\
 & 192 & 0.2167& \textbf{0.1857}& 0.2000& \textbf{0.1948}& 0.1941& \textbf{0.1914}& \textbf{0.1865}& 0.1903& 0.1992& 0.1995\\
 & 336 & 0.2641& \textbf{0.2396}& 0.2491& \textbf{0.2418}& 0.2500& \textbf{0.2476}& 0.2427& 0.2419& 0.2507& \textbf{0.2483}\\
\end{tabular}
}
\caption{Mean MSE of multi-variate time-series forecasting. Improvements of the hypernetwork over the base model and vice-versa are shown in \textbf{bold}.}
\label{table1}
\end{center}
\end{table*}

\begin{figure*}[t]
  \centering
  \includegraphics[width=\textwidth]{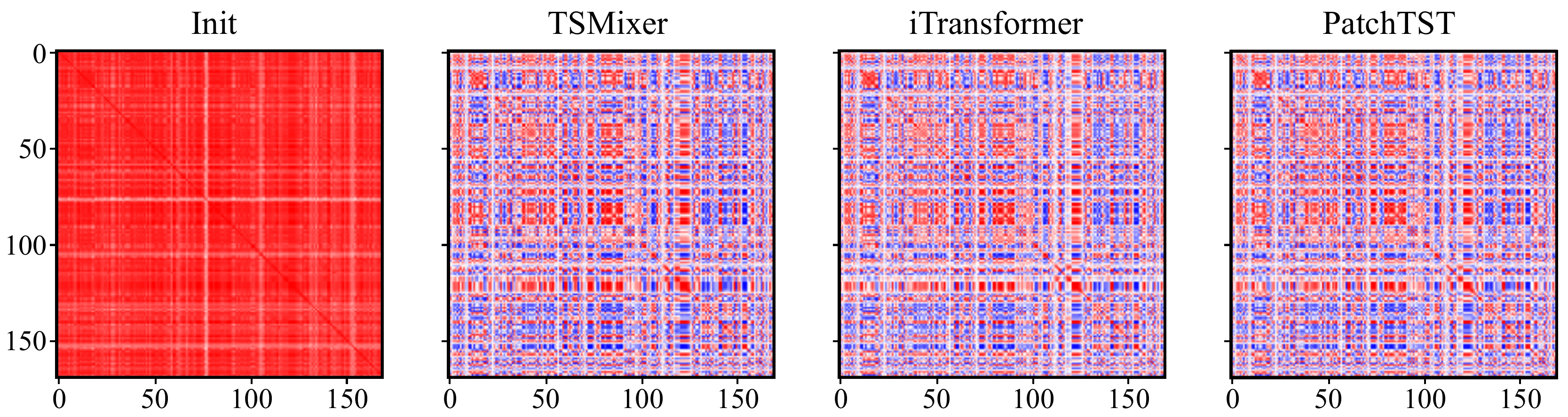}
  \caption{Embeddings for our HN-MVTS, PEMS08 dataset: (a) Initial embeddings before training, (b)-(d) learned embeddings for TSMixer, iTransformer and PatchTST.}\label{fig:hypernet_embeddings}
\end{figure*}

\begin{table*}
\begin{center}
\resizebox{\textwidth}{!}{
\begin{tabular}{l|l|lr|lr|lr|lr|lr}
Dataset & $H$ & DLinear & +HN-MVTS & TSMixer& +HN-MVTS & ModernTCN& +HN-MVTS   & PatchTST& +HN-MVTS & iTransformer& +HN-MVTS \\
\hline
ECL& 48 & 15.31$\pm$0.23 & 17.24$\pm$0.18 & 20.53$\pm$0.27 & 20.32$\pm$0.32 & 90.10$\pm$0.54 & 91.23$\pm$0.57 & 46.83$\pm$0.36 & 48.74$\pm$0.38 & 20.71$\pm$0.19 & 24.85$\pm$0.25 \\
 & 336 & 23.04$\pm$0.29 & 33.29$\pm$0.33 & 25.40$\pm$0.24 & 25.75$\pm$0.22 & 95.64$\pm$0.48 & 111.55$\pm$0.67 & 52.38$\pm$0.45 & 66.26$\pm$0.53 & 27.75$\pm$0.22 & 39.13$\pm$0.48 \\
\hline
ETTm1 & 48 & 4.71$\pm$0.01 & 5.61$\pm$0.05 & 2.87$\pm$0.02 & 3.36$\pm$0.04 & 15.87$\pm$0.09 & 11.01 $\pm$0.11& 8.71$\pm$0.08 & 11.37 $\pm$0.08& 7.18$\pm$0.12 & 9.13$\pm$0.11 \\
 & 336 & 5.39$\pm$0.05 & 8.12$\pm$0.03 & 3.06$\pm$0.07 & 4.76$\pm$0.06 & 10.22$\pm$0.10 & 17.27$\pm$0.13 & 6.16$\pm$0.04 & 12.90$\pm$0.08 & 7.87$\pm$0.03 & 10.82$\pm$0.14 \\
\hline
ETTm2 & 48 & 4.79$\pm$0.06 & 5.63 $\pm$0.07& 2.83$\pm$0.02 & 3.38$\pm$0.01 & 15.03$\pm$0.18 & 11.47$\pm$0.15 & 9.12$\pm$0.09 & 8.29$\pm$0.10 & 7.41$\pm$0.06 & 9.28$\pm$0.13 \\
 & 336 & 5.56$\pm$0.04 & 7.98$\pm$0.03 & 2.97$\pm$0.06 & 4.73$\pm$0.05 & 10.14$\pm$0.09 & 17.44$\pm$0.12 & 6.16$\pm$0.08 & 12.88$\pm$0.07 & 7.28$\pm$0.07 & 10.76$\pm$0.11 \\
\hline
PEMS03 & 48 & 16.67$\pm$0.16 & 18.83$\pm$0.13 & 19.42$\pm$0.19 & 20.13$\pm$0.17 & 99.71$\pm$0.39 & 102.28$\pm$0.50 & 51.58$\pm$0.28 & 54.14$\pm$0.34 & 22.14$\pm$0.23 & 27.26$\pm$0.26 \\
 & 336 & 25.31$\pm$0.28 & 35.84$\pm$0.40 & 26.57$\pm$0.37 & 26.17$\pm$0.34 & 107.65$\pm$0.54 & 121.93$\pm$0.62 & 58.46$\pm$0.29 & 73.27$\pm$0.61 & 29.60$\pm$0.28 & 43.39$\pm$0.31 \\
\hline
PEMS04 & 48 & 9.73$\pm$0.10 & 10.61$\pm$0.09 & 11.37$\pm$0.12 & 11.71$\pm$0.12 & 51.23$\pm$0.38 & 53.37$\pm$0.41 & 28.96$\pm$0.23 & 30.27$\pm$0.25 & 13.02$\pm$0.17 & 15.40$\pm$0.16 \\
 & 336 & 14.52$\pm$0.12 & 19.39$\pm$0.21 & 16.44$\pm$0.18 & 15.33$\pm$0.18 & 55.87$\pm$0.35 & 63.75$\pm$0.49 & 32.63$\pm$0.30 & 40.38$\pm$0.33 & 16.85$\pm$0.13 & 21.20$\pm$0.16 \\
\hline
PEMS07 & 48 & 40.87$\pm$0.34 & 44.62$\pm$0.35 & 33.85$\pm$0.39 & 34.86$\pm$0.33 & 298.28$\pm$0.78 & 300.66$\pm$0.82 & 133.25$\pm$0.54 & 139.22$\pm$0.61 & 57.74$\pm$0.47 & 77.60$\pm$0.85 \\
 & 336 & 59.21$\pm$0.37 & 83.22$\pm$0.44 & 53.67$\pm$0.41 & 58.45$\pm$0.50 & 315.42$\pm$0.92 & 372.10$\pm$1.05 & 153.61$\pm$0.64 & 188.44$\pm$0.77 & 84.84$\pm$0.59 & 107.37$\pm$0.63 \\
\hline
PEMS08 & 48 & 6.43$\pm$0.06 & 7.14$\pm$0.06 & 12.22$\pm$0.10 & 11.27$\pm$0.12 & 29.62$\pm$0.19 & 31.27$\pm$0.22 & 17.44$\pm$0.25 & 18.66$\pm$0.26 & 8.59$\pm$0.17 & 9.87$\pm$0.17 \\
 & 336 & 9.36$\pm$0.05 & 12.73$\pm$0.12 & 13.88$\pm$0.13 & 14.90$\pm$0.14 & 32.28$\pm$0.36 & 38.38$\pm$0.33 & 19.24$\pm$0.25 & 25.09$\pm$0.21 & 11.35$\pm$0.12 & 13.04$\pm$0.14 \\
\hline
Weather & 48 & 5.61$\pm$0.01 & 6.54$\pm$0.01 & 3.08$\pm$0.02 & 3.54$\pm$0.01 & 24.42$\pm$0.27 & 15.46$\pm$0.31 & 9.25$\pm$0.08 & 10.73$\pm$0.10 & 8.15$\pm$0.08 & 10.35$\pm$0.07 \\
 & 336 & 7.10$\pm$0.19 & 9.72$\pm$0.17 & 3.34$\pm$0.03 & 5.10$\pm$0.04 & 14.52$\pm$0.12 & 22.86$\pm$0.45 & 9.34$\pm$0.14 & 17.53$\pm$0.36 & 9.28$\pm$0.13 & 12.89$\pm$0.12 \\
\end{tabular}
}
\caption{Mean and standard deviation of training time (seconds) per epoch.
}
\label{table_times}
\end{center}
\end{table*}

\paragraph{Training and evaluation details.}
The training was performed on a server with 2x80Gb Nvidia A100 GPUs, Intel Xeon Gold 6326 CPUs (2.90GHz), and 512 GB of RAM. PyTorch 2.x was used to implement base models and our HN-MVTS. For the evaluation, we followed traditional settings~\cite{Nie2023,Wu2023}: we set the input length $T$ of the loopback window  to $336$, and evaluate the results for horizon lengths of $H = \{48, 96, 192, 336\}$. For a fair comparison, we apply reversible instance normalization \cite{Kim2021} for all models. We minimize the MSE with the Adam optimizer with a learning rate of 0.0001 and a batch size of 64. These hyperparameters enable us to achieve approximately the same results as those reported in the original papers~\cite{Zeng2023,Nie2023}. The dimensionality of embeddings for HN-MVTS is set to be less than or equal to the number of components in the time series. 

\section{Experimental Results}\label{sec:exper}
We report average MSE for four horizon lengths $H \in \{48, 96, 192, 336\}$ and 5 random seeds in Table~\ref{table1}. Here, the statistically significant (under the Wilcoxon signed-rank test with confidence 0.95 over 5 seeds) model among HN-MVTS and the baseline is shown in \textbf{bold}. The Mean Absolute Error (MAE) and detailed training curves, with the dependence of testing MSE on epoch, are presented in Appendix. 
As one can notice, in most cases, our HN-MVTS improves forecasting accuracy across all datasets and architectures, often by a substantial margin. 


Regarding applications to various architectures, the efficient and straightforward linear baseline, namely DLinear~\cite{Zeng2023}, benefits significantly from our method. For example, on the \texttt{Weather} dataset, MSE improves from 0.1369 to 0.1115 at horizon $H = 48$ and from 0.2641 to 0.2396 at $H = 336$. As a result, DLinear, combined with our HN-MVTS, became the top-performing solution for this dataset, although the vanilla DLinear was initially the worst model. Similar gains are observed on \texttt{ECL}, \texttt{PEMS04}, and \texttt{PEMS08}, i.e., improvements are dataset-dependent and often practically meaningful (particularly in high-dimensional or strongly correlated datasets). These results confirm that HN-MVTS can enhance lightweight models by incorporating channel relationships.

Second, as an MLP-based model, TSMixer~\cite{Chen2023} benefits from the added expressivity and channel specificity provided by our hypernetwork. For example, on the \texttt{ECL} dataset, MAE drops from 0.2453 to 0.2350 at $H = 48$, and from 0.3040 to 0.2876 at $H = 336$. Notably, TSMixer + HN-MVTS shows significant gains on \texttt{PEMS07} and \texttt{PEMS08}, which involve high-dimensional traffic sensor data with strong inter-channel dependencies. Third, while already competitive, ModernTCN~\cite{Luo2024} exhibits modest yet significant improvements when paired with HN-MVTS. On \texttt{ETTm2} and \texttt{PEMS} datasets, it becomes more robust at longer, typically more challenging horizons.

Finally, even advanced Transformer-based models benefit from our approach. On the \texttt{PEMS07} dataset at $H = 336$, PatchTST’s MSE decreases from 0.1619 to 0.1415, and iTransformer from 0.1030 to 0.0928. This demonstrates that our approach complements sophisticated attention mechanisms by generating adaptive, channel-aware output parameters. The visualization of HN-MVTS embeddings (Fig.~\ref{fig:hypernet_embeddings}) shows that channel embeddings learned for various base models are very similar and reflect the specifics of the dataset rather than the architecture of the MVTS model.

Regarding the dataset-specific observations, the \texttt{Weather} dataset shows the most significant improvement. For DLinear, the MSE at $H = 96$ drops from 0.1733 to 0.1425 (an 18\% relative improvement). 
\texttt{ECL} also sees noticeable improvement across horizons and architectures. For instance, TSMixer+HN-MVTS improves MSE from 0.1972 to 0.1666 at $H = 192$, showing that even a compact MLP benefits from our parameter-sharing scheme. Traffic datasets contain hundreds of correlated sensor channels, making them particularly suitable for channel-specific modeling. At $H = 336$, HN-MVTS improves PatchTST on \texttt{PEMS08} from 0.2343 to 0.2231, showing how our approach enhances scalability in high-dimensional input settings.

Thus, our experiments demonstrate that while the absolute gains vary across datasets and backbones, the improvements are consistent in direction and sometimes substantial (e.g., \texttt{Weather}, traffic datasets), reflecting HN-MVTS’s role as a general plug-and-play mechanism rather than a dataset-specific optimization. Our HN-MVTS improves all models across nearly every dataset and horizon length, confirming the value of channel-aware, dynamic parameter generation. Moreover, our approach is model-agnostic, yielding gains across all model types without requiring architectural overhauls, thereby underscoring the plug-and-play nature of our framework. Furthermore, we observe a resilience to horizon length $H$. Performance gains persist across short and long horizons. Particularly for longer-term forecasts (e.g., $H = 336$), where models typically degrade, HN-MVTS mitigates accuracy loss.

To assess the computational overhead introduced by HN-MVTS, we measured the average training time per epoch for each model on all datasets and forecast horizons. Results are presented in Table~\ref{table_times}. Despite introducing hypernetwork and learnable embeddings, the additional training cost remains modest. Across all models and datasets, the training time increase ranges from approximately 5\% to 25\%, depending on the model complexity and input size. The most lightweight models, such as DLinear and TSMixer, experience the most minor increases (e.g., DLinear on \texttt{ECL} at $H = 48$: from 15.31s to 17.24s, or \~12.6\%). The relative increase is moderate given their baseline costs, even for more complex Transformer-based models like PatchTST and iTransformer. Even on high-dimensional datasets (e.g., \texttt{PEMS07} with 883 channels), the added parameter cost is modest, and the increase in training time is negligible, demonstrating scalability. Thus, the HN-MVTS adds minimal training overhead, making it a practical enhancement for research and production scenarios. These efficiency gains and performance improvements further strengthen the case for integrating HN-MVTS into a wide range of forecasting pipelines.

\section{Conclusion}
In this work, we introduced HN-MVTS, a novel general-purpose framework for MVTS forecasting that combines the strengths of CI and CD modeling (Fig.~\ref{fig:forecasting_strategies}). Unlike traditional CI models, which ignore inter-channel relationships, or CD models, which can be overly complex, our method leverages the concept of hypernetworks to generate channel-specific parameters in a lightweight and scalable manner.  
We enable the model to condition its predictions on channel similarity, allowing shared statistical strength across similar components while still preserving the robustness and efficiency of CI models. HN-MVTS contributes a training-time, architecture-agnostic generative prior for per-channel final-layer weights. Importantly, our approach is modular and architecture-agnostic, allowing it to be seamlessly applied to a wide range of existing neural forecasting models (Fig.~\ref{fig:hypernet_architectures}), including both CI (DLinear, PatchTST) and CD (TSMixer, iTransformer). 
Through extensive empirical evaluations across multiple datasets, we demonstrated that HN-MVTS, in most cases, improves the accuracy of state-of-the-art forecasting models (Table~\ref{table1}). These gains come at minimal computational cost, with only a modest increase in training time and parameter count (Table~\ref{table_times}). The inference time is not affected, as the hypernetwork can be removed during the deployment phase after the model has been trained. 

Despite its flexibility and strong empirical results, HN-MVTS has several limitations. First, we assume that the last layer of a base forecasting model linearly transforms an input hidden state into the final output; therefore, multi-layer hyperparameterization (e.g., generating weights beyond the last layer) may offer deeper insights into representation sharing. Second, 
although the method is architecture-agnostic, integrating it into highly specialized, non-standard forecasting architectures may require additional tuning or design adaptation. We did not consider the application of hypernetworks to non-neural network models (gradient boosting, statistical models, etc.), which are still widely used.

Beyond immediate performance improvements, HN-MVTS opens up promising avenues for future research in adaptive and structured forecasting. Its ability to learn flexible representations of channel relationships makes it a natural fit for settings involving heterogeneous or evolving multivariate time series. Moreover, it provides a principled mechanism for incorporating prior knowledge, such as known similarity structures or hierarchical groupings, through initializing or designing component embeddings. 


\section{Acknowledgments}

The work was supported by the grant for research centers in the field of AI provided by the Ministry of Economic Development of the Russian Federation in accordance with the agreement 000000C313925P4E0002 and the agreement with HSE University No. 139-15-2025-009. 

The research was supported in part through the computational resources of HPC facilities at HSE University.

\bibliography{aaai2026_sup}

\clearpage
\onecolumn

\appendix
\section{Additional Experimental Results}\label{appendix:results}

\begin{table*}[h!]
\begin{center}
\resizebox{\textwidth}{!}{
\begin{tabular}{l|l|ll|ll|ll|ll|ll}
Dataset & $H$ & DLinear & +HN-MVTS & TSMixer& +HN-MVTS & ModernTCN& +HN-MVTS   & PatchTST& +HN-MVTS & iTransformer& +HN-MVTS \\
\hline
& 48 & 0.2231 & 0.2146 & 0.2453 & 0.2350 & 0.2339 & 0.2215 & 0.2125 & 0.2133 & 0.2058 & 0.2060 \\
ECL & 96 & 0.2365 & 0.2295 & 0.2794 & 0.2554 & 0.2552 & 0.2379 & 0.2276 & 0.2270 & 0.2263 & 0.2282 \\
 & 192 & 0.2495 & 0.2435 & 0.3005 & 0.2728 & 0.2773 & 0.2681 & 0.2442 & 0.2448 & 0.2507 & 0.2493 \\
 & 336 & 0.2652 & 0.2601 & 0.3040 & 0.2876 & 0.2813 & 0.2718 & 0.2606 & 0.2615 & 0.2638 & 0.2634 \\
\hline
& 48 & 0.3248 & 0.3199 & 0.3342 & 0.3376 & 0.3256 & 0.3258 & 0.3164 & 0.3259 & 0.3406 & 0.3461 \\
ETTm1 & 96 & 0.3436 & 0.3375 & 0.3563 & 0.3553 & 0.3473 & 0.3498 & 0.3440 & 0.3477 & 0.3557 & 0.3672 \\
 & 192 & 0.3637 & 0.3603 & 0.3786 & 0.3852 & 0.3753 & 0.3763 & 0.3653 & 0.3715 & 0.3770 & 0.3903 \\
 & 336 & 0.3841 & 0.3837 & 0.4030 & 0.3997 & 0.3999 & 0.4023 & 0.3921 & 0.3963 & 0.4005 & 0.4311 \\
\hline
& 48 & 0.2252 & 0.2214 & 0.2342 & 0.2332 & 0.2236 & 0.2196 & 0.2258 & 0.2269 & 0.2356 & 0.2367 \\
ETTm2 & 96 & 0.2521 & 0.2506 & 0.2652 & 0.2617 & 0.2546 & 0.2492 & 0.2566 & 0.2584 & 0.2685 & 0.2714 \\
 & 192 & 0.2893 & 0.2866 & 0.3120 & 0.3075 & 0.2972 & 0.2950 & 0.3025 & 0.3065 & 0.3164 & 0.3153 \\
 & 336 & 0.3253 & 0.3251 & 0.3447 & 0.3384 & 0.3361 & 0.3453 & 0.3399 & 0.3479 & 0.3506 & 0.3659 \\
\hline
& 48 & 0.2452 & 0.2452 & 0.2199 & 0.2248 & 0.2488 & 0.2325 & 0.2219 & 0.2173 & 0.1940 & 0.1942 \\
PEMS03 & 96 & 0.2705 & 0.2726 & 0.2430 & 0.2503 & 0.3118 & 0.3336 & 0.2477 & 0.2431 & 0.2161 & 0.2179 \\
 & 192 & 0.2825 & 0.2836 & 0.2569 & 0.2683 & 0.2968 & 0.2802 & 0.2690 & 0.2686 & 0.2307 & 0.2386 \\
 & 336 & 0.2965 & 0.2982 & 0.2721 & 0.2740 & 0.3400 & 0.3156 & 0.2787 & 0.2791 & 0.2457 & 0.2463 \\
\hline
& 48 & 0.2678 & 0.2608 & 0.2376 & 0.2377 & 0.2263 & 0.2238 & 0.2410 & 0.2258 & 0.1993 & 0.1949 \\
PEMS04 & 96 & 0.2928 & 0.2907 & 0.2483 & 0.2590 & 0.2905 & 0.2895 & 0.2676 & 0.2585 & 0.2189 & 0.2135 \\
 & 192 & 0.3088 & 0.3009 & 0.2748 & 0.2633 & 0.3017 & 0.2897 & 0.2848 & 0.2669 & 0.2306 & 0.2247 \\
 & 336 & 0.3235 & 0.3213 & 0.2939 & 0.2723 & 0.3205 & 0.3044 & 0.2977 & 0.2782 & 0.2539 & 0.2390 \\
\hline
& 48 & 0.2532 & 0.2499 & 0.2049 & 0.2137 & 0.2591 & 0.2339 & 0.2100 & 0.1949 & 0.1664 & 0.1618 \\
PEMS07 & 96 & 0.2854 & 0.2765 & 0.2264 & 0.2254 & 0.2478 & 0.2744 & 0.2342 & 0.2150 & 0.1796 & 0.1724 \\
 & 192 & 0.2984 & 0.2947 & 0.2489 & 0.2375 & 0.2570 & 0.2800 & 0.2539 & 0.2291 & 0.1907 & 0.1818 \\
 & 336 & 0.3185 & 0.3100 & 0.2754 & 0.2525 & 0.2590 & 0.2709 & 0.2670 & 0.2400 & 0.2023 & 0.1911 \\
\hline
& 48 & 0.2805 & 0.2718 & 0.2284 & 0.2223 & 0.2386 & 0.2449 & 0.2357 & 0.2194 & 0.1874 & 0.1902 \\
PEMS08 & 96 & 0.3186 & 0.3101 & 0.2629 & 0.2595 & 0.3570 & 0.3573 & 0.2643 & 0.2434 & 0.2069 & 0.1999 \\
 & 192 & 0.3424 & 0.3378 & 0.3035 & 0.2948 & 0.3310 & 0.2829 & 0.2947 & 0.2701 & 0.2222 & 0.2191 \\
 & 336 & 0.3601 & 0.3629 & 0.3168 & 0.3166 & 0.3090 & 0.3071 & 0.3058 & 0.2866 & 0.2371 & 0.2548 \\
\hline
& 48 & 0.1861 & 0.1514 & 0.1668 & 0.1650 & 0.1581 & 0.1563 & 0.1561 & 0.1576 & 0.1676 & 0.1761 \\
Weather & 96 & 0.2227 & 0.1894 & 0.2066 & 0.2037 & 0.1997 & 0.1955 & 0.1970 & 0.2010 & 0.2066 & 0.2037 \\
 & 192 & 0.2590 & 0.2315 & 0.2530 & 0.2443 & 0.2478 & 0.2379 & 0.2386 & 0.2446 & 0.2469 & 0.2545 \\
 & 336 & 0.2928 & 0.2733 & 0.2859 & 0.2835 & 0.2879 & 0.2858 & 0.2817 & 0.2863 & 0.2872 & 0.2905 \\
\end{tabular}
}
\end{center}
\caption{Mean MAE of multi-variate time-series forecasting.}

\label{table_mae}
\end{table*}

\begin{figure*}[t]
  \centering
  \includegraphics[width=0.85\textwidth]{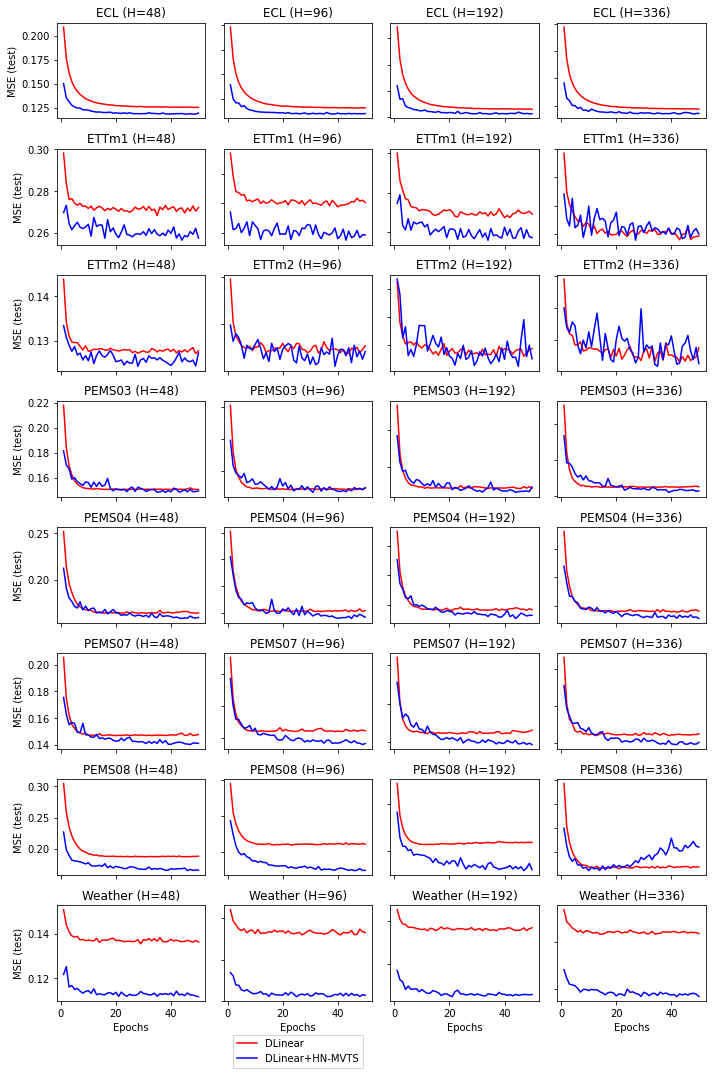}
  \caption{Training curves of MSE on a test set for DLinear base model.}\label{fig:DLinear_training_curves}
\end{figure*}

\begin{figure*}[t]
  \centering
  \includegraphics[width=0.85\textwidth]{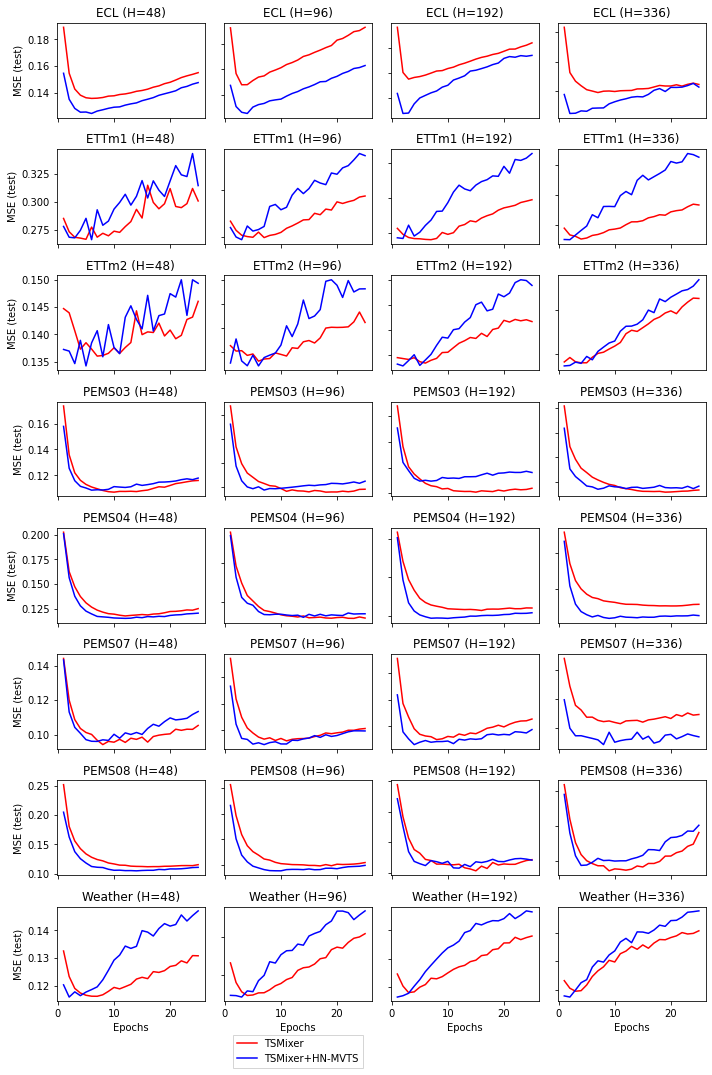}
  \caption{Training curves of MSE on a test set for TSMixer base model.}\label{fig:TSMixer_training_curves}
\end{figure*}

\begin{figure*}[t]
  \centering
  \includegraphics[width=0.65\textwidth]{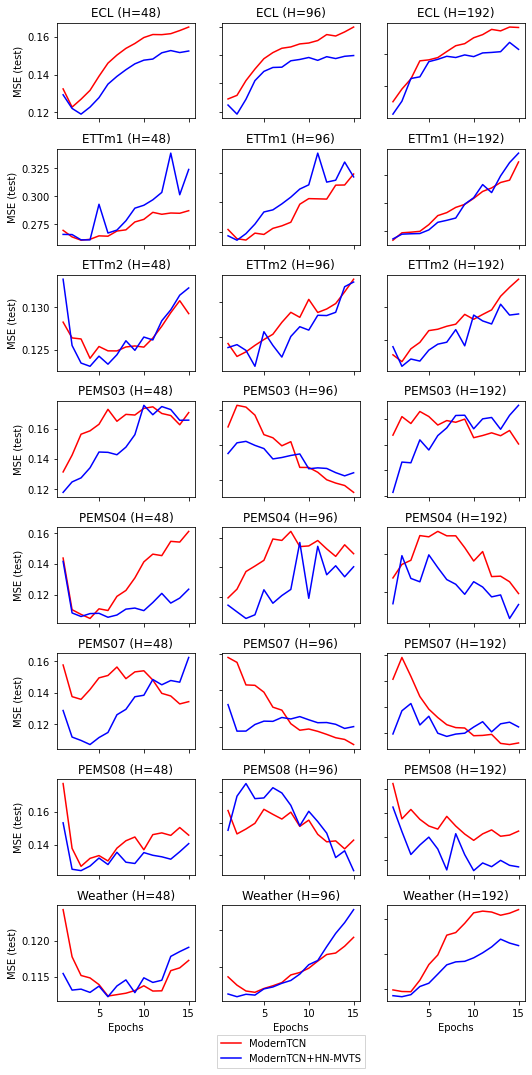}
  \caption{Training curves of MSE on a test set for ModernTCN base model.}\label{fig:ModernTCN_training_curves}
\end{figure*}

\begin{figure*}[t]
  \centering
  \includegraphics[width=0.85\textwidth]{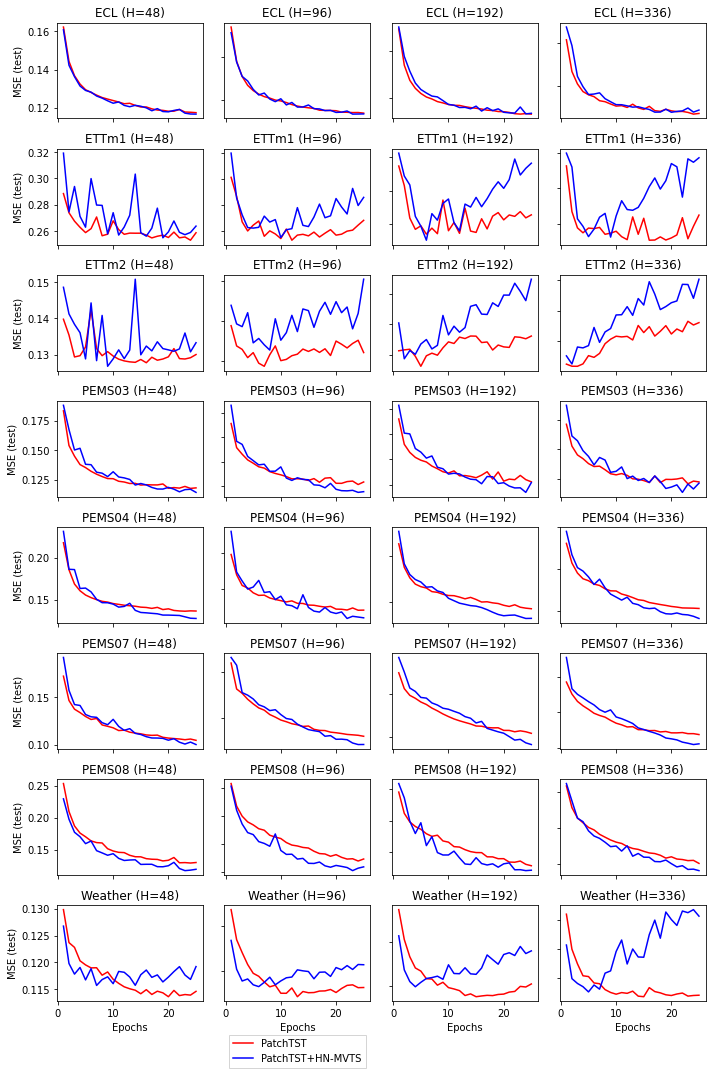}
  \caption{Training curves of MSE on a test set for PatchTST base model.}\label{fig:PatchTST_training_curves}
\end{figure*}

\begin{figure*}[t]
  \centering
  \includegraphics[width=0.85\textwidth]{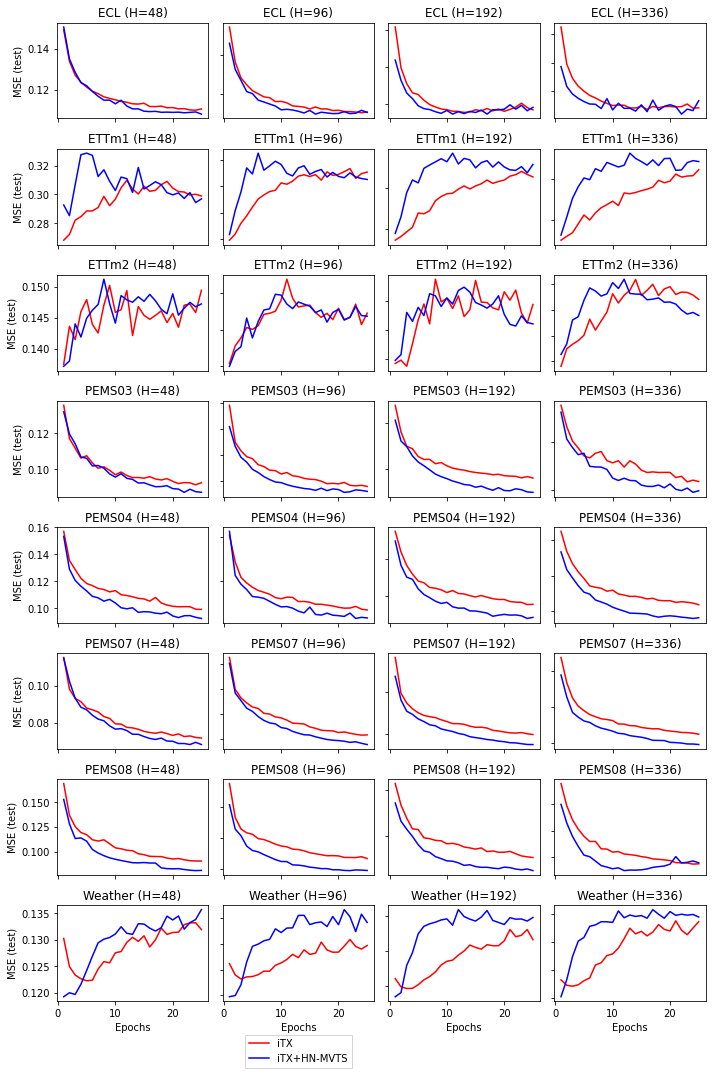}
  \caption{Training curves of MSE on a test set for iTransformer base model.}\label{fig:iTX_training_curves}
\end{figure*}

\end{document}